\documentclass[letterpaper]{article}
\usepackage{aaai22}
\usepackage{times}
\usepackage{helvet}
\usepackage{courier}
\usepackage{natbib}
\usepackage{amsmath}
\usepackage{amssymb}
\usepackage{graphicx}
\usepackage{footnote} 
\usepackage{hyperref} 

\begin{document}
%
\title{NumHTML: Numeric-Oriented Hierarchical Transformer Model \\ for Multi-task Financial Forecasting}

\author {
    Linyi Yang,\textsuperscript{\rm 1,2}
    Jiazheng Li,\textsuperscript{\rm 4}
    Ruihai Dong,\textsuperscript{\rm 3}
    Yue Zhang,\textsuperscript{\rm 1,2}
    Barry Smyth\textsuperscript{\rm 3}
}
\affiliations {
    \textsuperscript{\rm 1} Westlake Institute for Advanced Study, Westlake University\\
    \textsuperscript{\rm 2} School of Engineering, Westlake University\\
    \textsuperscript{\rm 3} The Insight Centre for Data Analytics, University College Dublin\\
    \textsuperscript{\rm 4} Department of Computer Science, University of Warwick\\
    {linyi.yang, yue.zhang}@westlake.edu.cn, {ruihai.dong, barry.smyth}@insight-centre.org, jiazheng.li@warwick.ac.uk
}

\maketitle
\begin{abstract}
\begin{quote}
Financial forecasting has been an important and active area of machine learning research because of the challenges it presents and the potential rewards that even minor improvements in prediction accuracy or forecasting may entail. Traditionally, financial forecasting has heavily relied on quantitative indicators and metrics derived from structured financial statements. Earnings conference call data, including text and audio, is an important source of unstructured data that has been used for various prediction tasks using deep earning and related approaches. However, current deep learning-based methods are limited in the way that they deal with numeric data; numbers are typically treated as plain-text tokens without taking advantage of their underlying numeric structure. This paper describes a numeric-oriented hierarchical transformer model (\emph{NumHTML}) to predict stock returns, and financial risk using multi-modal aligned earnings calls data by taking advantage of the different categories of numbers (monetary, temporal, percentages etc.) and their magnitude. We present the results of a comprehensive evaluation of NumHTML against several state-of-the-art baselines using a real-world publicly available dataset. The results indicate that NumHTML significantly outperforms the current state-of-the-art across a variety of evaluation metrics and that it has the potential to offer significant financial gains in a practical trading context. 
\end{quote}
\end{abstract}

\section{Introduction}

\noindent

It is the very nature of the stock market that even the most modest of advantages (e.g. speed of trade) can be parlayed into significant financial rewards, and thus traders have long been attracted to the idea of using historical data to predict future stock market trends. However, the stochastic nature of the stock market has proved to be very challenging when it comes to provide accurate future forecasts, especially when relying on pricing data alone \citep{moskowitz2012time,kristjanpoller2014volatility,manela2017volatility,zheng2019volatility,pitkajarvi2020cross}. However, recent advances in natural language processing (NLP) and deep learning (DL) introduce novels sources of data  --- textual data in the form of financial news \citep{Ding14,yang2018explainable,Liu2018WSDM,Liu2019KDD,du2020stock} and financial reports \citep{duan2018learning,Kogan09} to real-time social media \citep{Xu18,feng19} --- which may lead to be effective forecasting predictions. Of particular relevance to this paper is the earnings call data \citep{Kimbrough05,wang2014call,Qin19} that typically accompany the (quarterly) earnings reports of publicly traded companies. The multi-modal data associated with these reports include the textual data of the report itself plus the audio of the so-called earnings call where the report is presented to relevant parties, including a question-and-answer session with relevant company executives. The intuition is that the content of such a report and the nature of the presentation and Q\&A may encode valuable information to determine how a company may perform in the coming quarter and, more immediately relevant, how the market will respond to the earning report.

Previous work on using earnings conference calls has mostly considered the volatility prediction \citep{Qin19,yang2020html,sawhey2020mm,ye2020financial}, to predict the subsequent stock price fluctuation over a specified period (e.g., three days or seven days) after the earnings call \citep{bollerslev2016exploiting,Rekabsaz2017report}. An even more challenging, yet potentially more valuable signal, especially when it comes to optimizing a real-world trading strategy, is the predicted \emph{stock return}. While this has been explored by using financial news data \citep{Ding14,Ding15,chang2016measuring,duan2018learning,yang2019leveraging,du2020stock} and analyst reports \citep{Kogan09,loughran2011liability,chen2021opinion}, the use of earnings call data remains largely unexplored. One notable exception is \citet{sawhey2020mm}, which considers stock movement prediction as an auxiliary task for enhancing the financial risk predictions. In particular, they show that multi-task learning is useful for improving volatility prediction at the expense of lower price movement prediction accuracy. 

The main objective of this work is to explore the use of earnings calls data for stock movement prediction. The starting point for this work is the multi-task learning approach described by \citep{yang2020html}, which leverages textual and audio earnings call data with additional vocal features extracted by Praat \citep{boersma2001speak}. We quantify the effectiveness of textual and vocal information from earnings calls to predict stock movement using this baseline model and then go on to extend this model \cite{yang2020html} by describing three auxiliary loss functions to investigate the utility of a more sophisticated representation of numerical data during prediction. Thus the central advance in this work is a novel, numeric-oriented hierarchical transformer model (\emph{NumHTML}) for the prediction of stock returns, using multi-modal aligned earnings calls data by taking advantage of the auxiliary task (volatility prediction), different categories of numerical data (monetary, temporal, percentages etc.), and their magnitude, motivated by the fact that volatility is a relevant factor to future stock trends and the assumption that better numerical understanding can benefit forecasting. These components are integrated through a novel structured adaptive pre-training strategy and Pareto Multi-task Learning.

We present the results of a comprehensive evaluation on a real-world earnings call dataset to show how the model can be more effective than the baseline system, facilitating more accurate stock returns predictions without compromising volatility prediction \citep{Qin19,yang2020html,sawhey2020mm}. Also, the results of a realistic trading simulation shows how our approach can generate a significant arbitrage profit using a real-world trading dataset. All code and datasets will be released on GitHub.

\section{Related Work}
This paper brings together several areas of related work -- \textit{Stock Movement Predictions}, \textit{Multi-modal Aligned Earnings Call}, and \textit{Representing Numbers in Language Models} -- and in what follows, we briefly summarise the relevant state-of-the-art in each of these areas as it relates to our approach. We are the first to examine whether pre-trained models with better numerical understanding can improve performance on financial forecasting tasks based on the multi-modal data.

\subsection{Stock Movement Prediction}
While there has been a long-standing effort when it comes to applying machine learning techniques to financial prediction \citep{da2015sum,xing2018intelligent,xing2019sentiment}, typically by using time-series pricing data, reliable and robust predictions have proven to be challenging due to the stochastic nature of stock markets. However, recent work has shown some promise when it comes to predicting stock price movements using deep neural networks with rich textual information from financial news and social media (primarily Twitter) \citep{liu2013estimation,Ding14,Ding15,Xu18,duan2018learning,yang2018explainable,feng19}. By taking advantage of much richer sources of relevant data (news reports, export commentaries etc.), deep learning techniques have been able to generate more robust and accurate predictions even in the face of market volatility.

Elsewhere, researchers have considered the role of opinions in financial prediction. For example, one recent study \citep{chen2021evaluating} has shown how that the opinions from company executives and managers or financial analysts can be more effective than the opinions of amateur investors when it comes to predicting the stock price. However, previous works using earnings calls data typically focus on the financial risk (volatility) prediction, while whether volatility prediction can help predict movement has been less well covered.

\subsection{Representing Numbers in Language Models}
Current language models treat numbers within the text input as plain words without understanding the basic numeric concepts. Given the ubiquity of numbers, their importance in financial datasets, and their fundamental difference with words, developing richer representations of numbers could improve the model's performance in downstream financial applications \citep{chen2019numeracy,sawhey2020mm}. Progress towards deeper numerical representations has been limited but promising. For example, previous work, represented by the DROP \citep{dua2019drop}, presents a variety of numerical reasoning problems. Different from the existing works that pay attention to explore the capability of pretrained language models for general common-sense reasoning \cite{zhang2020language}, and math word problem-solving \cite{wu2021math}, we focus on improving the numeral understanding ability of language models for financial forecasting, motivated by the fact that financial documents often contain massive amounts of numbers. In particular, we consider two tasks -- Numeral Category Classification \citep{chen2018cikm,chen2019numeracy,chen2021evaluating} and Magnitude Comparison \citep{wallace2019nlp,naik2019exploring} -- using a structured adaptive pre-training strategy to improve the capability of pretrained language models for multi-task financial forecasting.

\subsection{Multi-modal Aligned Earnings Call Data}
Earnings conference call is typically presented by leading executives of publicly listed companies and provide an opportunity for the company to present an explanation of its quarterly results, guidance for the upcoming quarter, and an opportunity for some in-depth Q\&A between the audience and company management \citep{Keith19}. An early study \citep{larcker2012detecting} mentioned that text-based models could reveal misleading information during earnings calls and cause stock price swings in financial markets and most unstructured data resources are still text data. So far, the multi-modal aligned earnings call data mainly refers to the sentence-level text-audio paired data resource, represented by \citet{Qin19} and \citet{li2020maec}. While previous works \cite{Qin19,yang2020html,sawhey2020mm} mainly explore the benefits of audio features for volatility predictions, we propose a different research question that whether adaptive pre-training and volatility prediction loss can benefit the performance of stock prediction by using multi-modal aligned earnings call data.


\begin{figure*}[ht]
  \centering
  \includegraphics[width = .68\linewidth]{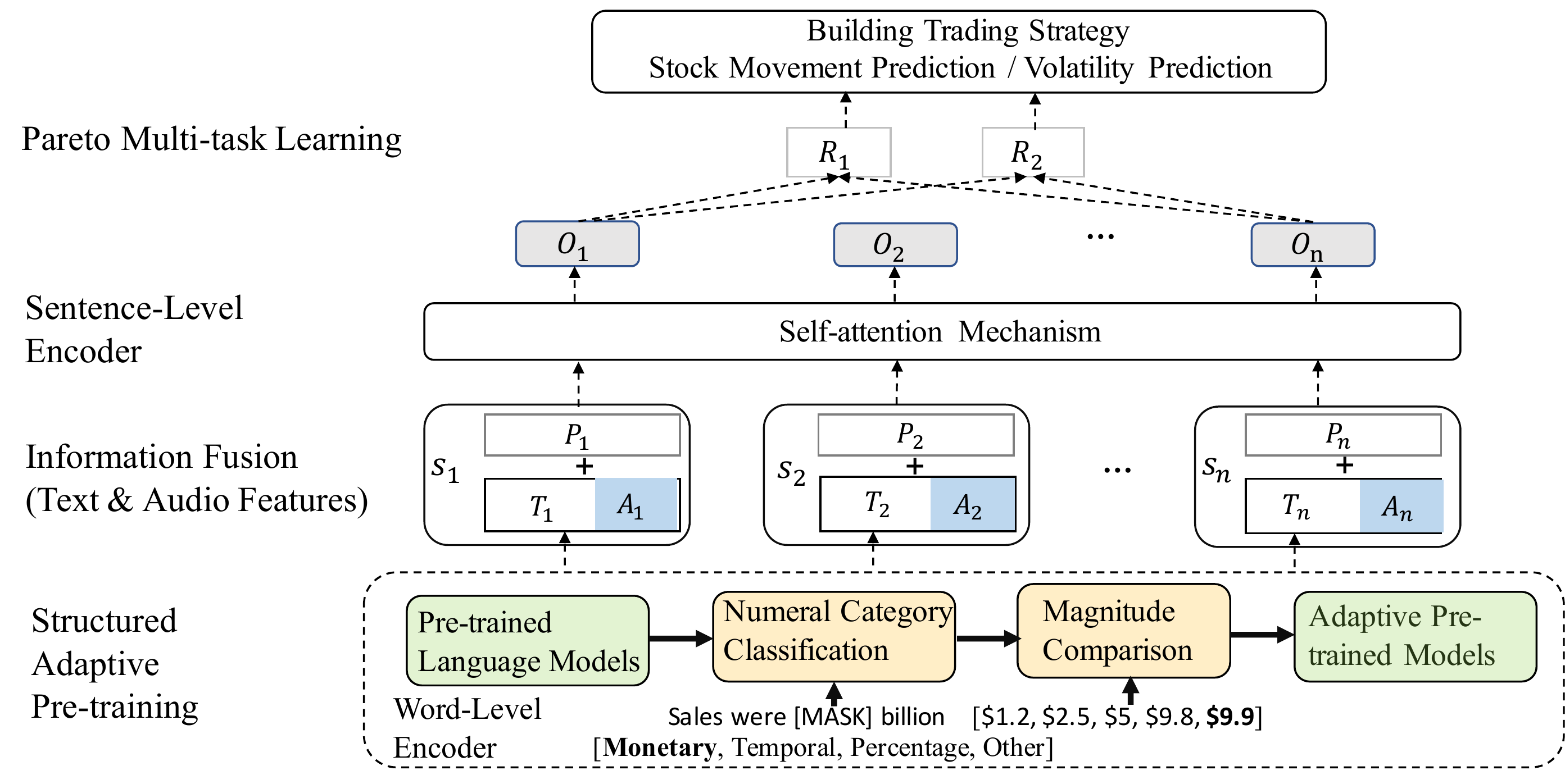}
  \caption{Overall architecture of NumHTML. We use the surrounding tokens around [MASK] to classify the numeral categories.}
  \label{fig:c61}
\end{figure*}

\section{Approach}

The NumHTML model proposed in this paper is shown in Figure \ref{fig:c61} and is made up of four components: (1) \emph{word-level encoder}; (2) \emph{multimedia information fusion}; (3) \emph{sentence-level encoder}; and (4) \emph{pareto multi-task learning}. Briefly, a key innovation in this work is the use of a novel structured adaptive pre-training approach to improve how numeric information is treated during the word-level encoding of earnings calls transcripts. Then, sentence-level text features are aligned with 27 classical audio features extracted from earnings calls audio \citep{boersma2001speak} based on the approach described by \citet{Qin19}. Next, the information fusion layer is responsible for combining the resulting text and audio features into a single representation for use by the sentence-level encoder to generate a multi-modal input that is suitable for training and prediction.




\subsection{Structured Adaptive Pre-training}
The pre-training process consists of two main tasks,  \textit{Numeral Category Classification} and \textit{Magnitude Comparison}, in order to improve the representation of numerical data. During \textit{Numeral Category Classification}, sentences that contain numeric data are categorised as belong to one or more of four main classes: \emph{monetary}, \emph{temporal}, \emph{percentage}, and \emph{other}. This categorization process uses a set of \emph{trigger tokens} and rules so that, for example, in the sentence "During 2020 profits increased by 13\% to \$205m" presumably this is tagged as monetary (because of the \$205m), temporal (2020) and percentage (\%). We freeze the penultimate layer of fine-tuned whole-word-masked BERT (WWM-BERT) model before fine-tuning for the next numeral understanding task, \textit{Magnitude Comparison}.



Following \citet{wallace2019nlp}, \textit{Magnitude Comparison} is probed in an argmax setting. Given the embeddings for five numbers, the task is to predict the index of the maximum number. Each list consists of values of similar magnitude within the same numeral type in order to conduct a fine-grained comparison. For example, for a given list containing five monetary numbers \([\$1.2, \$2.5, \$5, \$9.8, \$9.9]\), the training goal is to find the largest position value within this five values. In this given example, the golden label should be \([0,0,0,0,1]\). Softmax is used to assign a probability to each index using the hidden state trained by the negative log-likelihood loss. In practice, we shape the training/test set by uniformly sampling raw earnings call transcript data without placing back. A BiLSTM network will be fed with the list of token embeddings -- varying from the pre-trained embeddings to the fine-tuned embeddings -- connected with a weight matrix to compare its performance. The token-level encoder tuned by the structured adaptive pre-training is used for shaping the sentence-level textual embedding.

\subsection{Sentence-level Transformer Encoder}
We adopt a sentence-level Transformer encoder fed with sentence-level representations of long-form multi-modal aligned earnings call data (usually contains more than 512 tokens) for multi-task financial forecasting. Let \(W_{i} =  \left(w_{i}^{1}, w_{i}^{2},...,w_{i }^{\left|W_{i}\right|}\right)\) be a sentence, where \(|W_i|\) is the length and \(w_{i}^{\left|W_{i}\right|}\) is an artificial \verb|EOS| (end of sentence) token. The word embedding matrix associated with \(W_i\) is initialized as
\begin{equation}
\begin{aligned} \mathbf{E}_{i}=&\left(\mathbf{e}_{i}^{1}, \mathbf{e}_{i}^{2}, \ldots, \mathbf{e}_{i}^{\left|t_{i}\right|}\right) \\ \text { where } & \mathbf{e}_{i}^{j}=e\left(w_{i}^{j}\right)+\mathbf{p}_{j} .\end{aligned}
\end{equation} \(e(\cdot)\) maps each token to a \(d\) dimensional vector using WWM-BERT, and \(p_j\) is the position embedding of \(w_i^{j}\) with the same dimension \(d\). Consequently, \(e_i^{j} \in \mathbb{R}^{d}\) for all \(j\). 

The enhanced WWM-BERT model after structured adaptive pre-training is adopted as the token-level Transformer encoder. The sentence representation \({T_{i}} \in \mathbb{R}^{d_{t}} \) of the sentence \(W_{i}\) is calculated through the average pooling operating over the second last layer of the network. $d_{t}$ represents the default dimensions of word embeddings. The sentence representations are aligned with sentence-level audio features in the information fusion layer later. Finally, the multi-modal representation of a single earnings call is represented as:

\begin{equation}\begin{aligned}
\mathcal{D}^{(k)}=&\left(s_{1}^{(k)}, s_{2}^{(k)}, \ldots, s^{(k)}_{M}\right)  \\ \text { where } & \mathbf{s}_{i}^{(k)}=\left((T_{i}^{(k)},A_{i}^{(k)})+P_{i}\right).\end{aligned}
\end{equation}

\(T_{i}^{k}\) and \(A_{i}^{k}\) represent the textual and audio features of sentence \(i\) of document \(\mathcal{D}^{(k)} \in \mathbb{R}^{M \times d_{s}} \), respectively, and \(P_{i} \in \mathbb{R}^{M \times d_{s}} \) denotes the trainable sentence-level position embedding. \(M\) is the maximum number of sentences. 

\subsection{Pareto Multi-task Learning}
We adopt the Pareto Multi-task Learning algorithm (Pareto MTL) proposed by \citet{lin2019pareto} to integrate stock movement prediction and volatility prediction by finding a set of well-distributed Pareto solutions that can represent different trade-offs between both tasks. Pareto MTL decomposes a Multi-Task Learning (MTL) problem into multi-objective subproblems with multiple constraints. An average pooling operation is first applied to the output of the sentence-level Transformer encoder. Then, we find a set of well-distributed unit preference vectors \(\left\{u_{1}, u_{2}, \ldots, u_{K}\right\}\) in \(R_{+}^{2} \); $K=10$ in this work. The multi-objective sub-problem corresponding to the preference vector \(u_{K}\) is defined as:
\begin{equation}
\begin{array}{c}\min _{\theta} \mathcal{L}(\theta)=\left(\mathcal{L}_{1}(\theta), \mathcal{L}_{2}(\theta), \cdots, \mathcal{L}_{m}(\theta)\right)^{\mathrm{T}}, \text {s.t.} \mathcal{L}(\theta) \in \Omega_{k} \end{array}\\ 
\end{equation} 
The idea of Pareto MTL is shown in Figure 2, where \(\mathcal{L}_{m}(\theta)\) is the loss of task \(m\) and \(\Omega_{k}\)(k=1,\dots, K) is a sub-region in a objective space and with \(u_{j}^{T} v\) as the inner product between the preference vector \(u_{j}\) and a given vector \(v\):
\begin{equation}
\begin{array}{c}\Omega_{k}=\left\{\boldsymbol{v} \in R_{+}^{2} \mid \boldsymbol{u}_{j}^{T} \boldsymbol{v} \leq \boldsymbol{u}_{k}^{T} \boldsymbol{v}, \forall j=1, \dots, K\right.\}\end{array}
\end{equation}  Hence, the set of possible solutions in different sub-regions represent different trade-offs among these two tasks. A two-step, gradient-based method is used to solve these multi-objective sub-problems based on the sentence-level multi-modal representations. 

\begin{figure}[t]
  \centering
  \includegraphics[width = .65\linewidth]{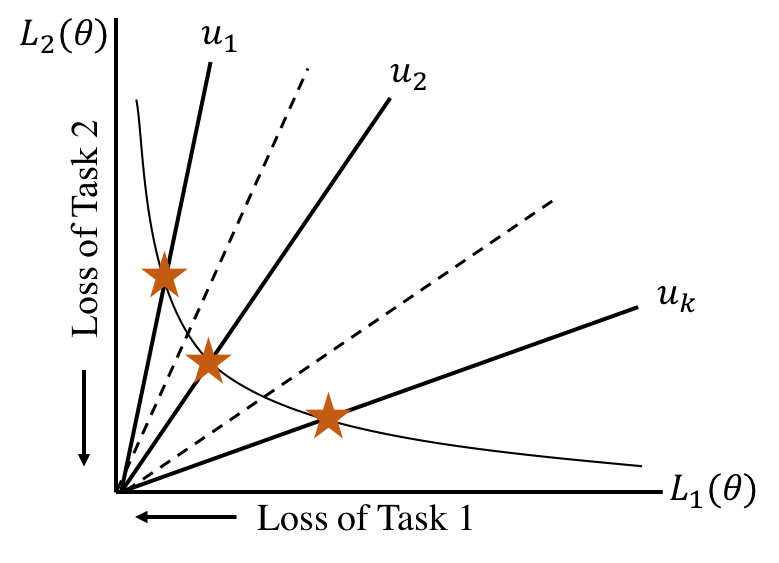}
  \caption{Pareto MTL aims to find a set of Pareto solutions in different restricted preference regions \citep{lin2019pareto}.}
  \label{fig:c62}
\end{figure}

\paragraph{Initial Solution:}
To find an initial feasible solution \(\theta_{0}\) for a high-dimension, constrained optimization problem, we use a sequential gradient-based method since the straightforward projection approach is too expensive to calculate directly for the 345-million parameter WWM-BERT model. The update rule used is \(\theta_{t+1}=\theta_{t}+\eta d_{t}\) where \(\eta\) is the step size and \(d_{t}\) is the search direction, which can be obtained from the rule of Pareto critical \citep{zitzler1999multiobjective}. Iteration terminates once a feasible solution is found or the max number of iterations is met. 


\paragraph{Achieving Pareto Efficiency:}
The next step is to solve the constrained subproblems in order to find a set of distributed solutions that can achieve the Pareto efficiency. Following \citet{lin2019pareto}, we obtain a restricted Pareto critical solution for each training goal by using constrained multi-objective optimization, which generalizes the steepest descent method for unconstrained multi-objective optimization problems. Due to the high-dimensionality of the problem, we change the decision space from the parameter space to a more tractable objective and constraint space; see \citet{lin2019pareto} for a proof of this and the algorithm used. The result is a reduction in the dimension of the optimization problem from 345 millions to seven (two objective functions plus five activated constraints), which allows the Pareto MTL to be scaled and optimized for our task as shown in Equation 5, where \(\hat{y}_{i}\) and \(\hat{y}_{j}\) are the predicted values for the main and auxiliary tasks, respectively, and \(y_{i}\) and \(y_{j}\) denote the corresponding true values. The output of Pareto MTL is a set of weight allocation strategies (\(\alpha_{patero_{1}}\) and \(\alpha_{patero_{1}}\)) for both tasks. We use Adam \citep{kingma2014adam} as the optimizer and adopt the trick of learning-rate decay with increasing steps to train the model until it converges.


\begin{equation}
\mathcal{F} = \alpha_{patero_{1}} \sum_{i}\left(\hat{y}_{i}-y_{i}\right)^{2} + \alpha_{patero_{2}} \sum_{j}\left(\hat{y}_{j}-y_{j}\right)^{2}
\end{equation}

\section{Evaluation}
We make a comprehensive comparison of NumHTML with several state-of-the-art baselines using a publicly available dataset, by first focusing on stock movement prediction (Section \ref{sec:stock}) and then by testing various stock prediction techniques in a realistic long-term trading simulation (Section \ref{sec:trade}). The hyper-parameters for our method and baselines are all selected by a grid search on the validation set. In each case, we demonstrate the significant advantage of NumHTML compared to baselines. Prior to these studies, we describe the intermediate results for the adaptive training used by NumHTML to demonstrate how it significantly outperforms the conventional pre-trained model.


\subsection{Dataset \& Methodology}
\subsubsection{Dataset:} In line with previous work \citep{yang2020html,sawhey2020mm} for multi-task financial forecasting, in this evaluation, we use a publicly available Earning Conference Calls dataset constructed by \citet{Qin19}. This dataset contains 576 earning calls recordings, correspond to 88,829 text-audio aligned sentences, for S\&P 500 companies in U.S. stock exchanges. The dataset also includes the corresponding dividend-adjusted closing prices from Yahoo Finance \footnote{\url{https://finance.yahoo.com/}} for calculating volatility and stock returns. To facilitate a direct comparison with the current state-of-the-art \citep{sawhey2020mm}, we split the dataset into mutually exclusive training/validation/testing sets in the ratio of 7:1:2 (refers to instances) in chronological order, since future data cannot be used for prediction.

\subsubsection{The Stock Prediction Task:} We evaluate the stock prediction task as a classification problem --- that is, the task is to predict whether a stock moves up (positive) or down (negative) due to the earnings call --- in order to ensure a fair comparison with \citep{sawhey2020mm}. The prediction of \emph{n-day} stock movement will be a rise if the regression results of the stock return is a positive value and vice versa. 

\subsubsection{The Trading Simulation Task:} To perform a trading simulation based on the multi-modal multi-task learning architecture, we aim to optimize for (1) average \emph{n-day} volatility (that is, the average volatility of the following $n$ days); and (2) cumulative \emph{n-day} stock return (that is, the cumulative profit $n$ days after an earnings call). In the trading simulation, stock movement predictions are used to decide whether to buy or sell a stock after $n$ days. To ensure a fair comparison, we use the trading strategy implemented by \citep{sawhey2020mm}, which relies on the results of stock movement prediction when \(n = 3\). Thus, if the prediction is a rise in price \(p_{d-n}^{s}\) from day \(d-n\) to \(d\) for stock \(s\), the strategy buys the stock \(s\) on day \(d-n\), and then sells it on day \(d\).  In addition, we perform a short sell \footnote{Short sell: \url{https://en.wikipedia.org/wiki/Short_(finance)}} if the prediction is a fall in price. We maintain the same trading environment with \citet{sawhey2020mm}: there are no transaction fees, we can only purchase a single share (but for multiple companies) for each time period, and intra-day trading is not considered \footnote{Obviously, this represents a simplified trading strategy, given that it is limited to single share purchases. It was adopted here to align with previous work \citep{sawhey2020mm} but is an obvious avenue for future work to implement more sophisticated trading strategies.}. 


\subsection{Evaluation Metrics}
For stock movement prediction we report the F1 score and Mathew's Correlation Coefficient (MCC) for stock price prediction. MCC performs more precisely when the data is skewed by accounting for the true negatives. For a given confusion matrix:

\begin{equation}
MCC=\frac{t p \times t n-f p \times f n}{\sqrt{(t p+f p)(t p+f n)(t n+f p)(t n+f n)}}
\end{equation}

Then the predicted average \(n-day\) volatility is compared with the actual volatility (Eq. 7) to compute the mean squared error for each hold period: \(n \in\{3,7,15,30\}\). 
\begin{equation}
v_{[0, n]}=\ln \left(\sqrt{ \left.\frac{\sum_{i=1}^{n}\left(r_{i}-\bar{r}\right)^{2}}{n}\right)}\right.
\end{equation}
\(r_{i}\) is the stock return on day \(i\) and \(\bar{r}\) is the average stock return (using adjusted closing price) in a window of \(n\) days.

\begin{equation}
\label{eq:mse_}
MSE=\frac{\sum_{i}\left(\hat{y}_{i}-y_{i}\right)^{2}}{n}
\end{equation}

For the stock trading simulation we use the cumulative profit and Sharpe Ratio metrics. The cumulative profit generated by a simple trading strategy is defined as:
\begin{equation}
\text {Profit}=\sum_{s \in S}\left(p_{d}^{s}-p_{d-\tau}^{s}\right) *(-1)^{\text {Action }_{s}^{d-\tau}}
\end{equation}
where \(\left(p_{d}^{s}\right)\) indicates the stock price of stock \(s\) on the day \(d\), and Action \(_{s}^{d-\tau}\) is a binary value depended on the stock movement prediction result; it equals to 0 if the model predicts a rise in price for stock \(s\) on day \(d\), otherwise it is \(1\). Sharpe Ratio evaluates the performance of investments using their average return rate \(r_{x}\), risk-free return rate \(R_{f}\) and the standard deviation \(\sigma\) across the investment \(x\):
\begin{equation}
\text {Sharpe Ratio}=\frac{r_{x}-R_{f}}{\sigma\left(r_{x}\right)}
\end{equation}

\begin{table}[t]
\small
\centering 
\caption{The four-class numeral category classification results varying from different embeddings.}
\begin{tabular}{lccc}
\hline
\textbf{Model} & \textbf{LRAP} & \textbf{ROC\_AUC} \\ \hline
Glove & 0.870 & 0.858 \\
WWM-BERT & 0.920 & 0.904 \\
WWM-BERT+NCC & 0.973 & 0.977\\ \hline
\end{tabular}
\end{table}

\begin{table*}[t]
\small
\centering 
\caption{The Magnitude Comparison Results (List Maximum from 5-numbers).}
\begin{tabular}{lcccc}
\hline
\textbf{Model} & \textbf{Monetary} & \textbf{Temporal} & \textbf{Percentage} & \textbf{All} \\ \hline
GloVe & 0.84 & 0.78 & 0.89 & 0.82 \\
WWM-BERT & 0.90 & 0.71 & 0.95 & 0.88  \\
WWM-BERT+NCC & 0.89 & 0.72 & 0.95 & 0.88 \\ 
WWM-BERT+NCC+MC & 0.93& 0.85 & 0.99 & 0.94 \\ \hline
\end{tabular}
\end{table*}

\subsection{Baselines}

We consider several different baselines \citep{wang2016attention,yang2016hierarchical,Qin19,yang2020html,sawhey2020mm}, which, to the best of our knowledge, offer the best available stock prediction methods at the time of writing. These baselines can be grouped according to whether they use historical (numeric) pricing data, textual earnings call data, or multi-modal earnings call data. 

\begin{enumerate}

    \item {\bfseries LSTM+ATT \citep{wang2016attention}:} The best performing price-based model (LSTM with attention) in which the n-day volatility in the training data is predicted using pricing data only.
    
    \item {\bfseries HAN (Glove) \citep{yang2016hierarchical}:} Uses textual data in which each word in a sentence is converted to a word embedding using the pre-trained Glove 300-dimensional embeddings and trained by a hierarchical Bi-GRU models \citep{bahdanau2014neural}.
    
    \item {\bfseries MDRM \citep{Qin19}:} This recent work was the first to consider  volatility prediction a multi-modal deep regression problem based on a newly proposed multi-modal aligned earnings calls dataset. 
    
    \item {\bfseries HTML \citep{yang2020html}:} This recent hierarchical transformer-based, multi-task learning framework is designed specifically for volatility prediction using multi-modal aligned earnings call data. 
    
    \item {\bfseries Multi-Modal Ensemble Method \citep{sawhey2020mm}:} This multi-modal, multi-task learning approach represents the current state-of-the-art in the task of stock movement predictions using a combination of textual and audio earnings calls data.
\end{enumerate} 

\subsection{Evaluating Structured Adaptive Training} 
To begin with, we present the results on the validation set for the adaptive training used by NumHTML. 

\subsubsection{Numeral Category Classification}
The results of multi-label numeral category classification (NCC) on the validation set are shown in Table 1. The aim is to show how this task significantly enhances the token-level embeddings. Both Label ranking average precision (LRAP) and ROC\_AUC scores of the financial numeral category classification have been increased with the benefit of adaptive pre-training, which suggests that our approach (BERT+NCC) can classify numeral categories better than the raw pre-trained embeddings, including BERT and Glove. In particular, the performance of the adaptive pre-trained model is improved around 5.3\% in LRAP and 7.3\% in ROC\_AUC. 

\subsubsection{Magnitude Comparison}
The accuracy of the Magnitude Comparison (listed maximum 5-numbers) based on different methods are shown in Table 2. We find that the `NCC' task cannot guarantee the accuracy benefits for the maximum list task. However, the adaptive pre-training directly on the magnitude comparison task can significantly improve performance (94\% vs 88\% on average). We also notice that the 'percentage' classification can achieve the highest accuracy among four types, while the secular values are the hardest to predict (85\% vs 99\%). We speculate that numbers representing percentages are between 0 and 99, making it easier to predict the largest number among them. On the other hand, the numbers representing years usually contain four digits and are similar, posing a challenge for the magnitude comparison. 

\subsection{Evaluating Stock Movement Prediction \label{sec:stock}}

\begin{table*}[t]
\small
\centering 
\caption{Results for the future n-day stock movement prediction (higher is better). * and ** indicate statistically significant improvements over the state-of-the-art ensemble method with p\(<\)0.05, p\(<\)0.01 respectively, under Wilcoxon’s test.}
\begin{tabular}{l|llll|llll}
\hline
 & \multicolumn{8}{c}{Price Movement Predictions} \\ 
Model & \(MCC_3\) & \(MCC_7\) & \(MCC_{15}\) & \(MCC_{30}\) & \(F1_3\) & \(F1_7\) & \(F1_{15}\) & \(F1_{30}\) \\ \hline
Price-based LSTM & 0.069 & 0 & 0.097 & 0 & 0.271 & 0.694 & 0.200 & 0.765 \\
Price-based BiLSTM-ATT & 0 & 0 & 0 & 0 & 0.149 & 0.342 & 0.200 & 0.721 \\ \hline
SVM & -0.069 & 0.015 & -0.048 & -0.003 & 0.524 & 0.683 & 0.645 & 0.734 \\
HAN (Glove)& 0.090 & -0.005 & 0.266 & -0.042 & 0.591 & 0.621 & 0.598 & 0.703 \\ \hline
\multicolumn{9}{l}{\textbf{Text-only Methods}} \\ \hline
MDRM & 0.117 & -0.107 & 0.032 & -0.085 & 0.675 & 0.500 & 0.571 & 0.601 \\
HTML & 0.195 & 0.007 & 0.119 & 0.022 & 0.623 & 0.688 & 0.648 & 0.700 \\
Ensemble & 0.204 & 0.008 & \textbf{0.132} & 0.024 & 0.675 & 0.690 & 0.636 & 0.703 \\
NumHTML & \textbf{0.229**} & \textbf{0.009} & 0.122 & \textbf{0.031} & \textbf{0.689*} & \textbf{0.691} & \textbf{0.644**} & \textbf{0.727*} \\\hline 

\multicolumn{9}{l}{\textbf{Multi-modal Methods}} \\ \hline
MDRM (Multi-modal) & 0.095 & 0.056 & 0.159 & -0.065 & 0.628 & 0.690 & 0.452 & 0.590 \\ 
HTML (Multi-modal) & 0.280 & 0.125 & 0.196 & 0.131 & 0.696 & 0.695 & 0.703 & 0.748 \\ 
Ensemble (Multi-modal) & 0.321 & 0.128 & 0.191 & 0.128 & 0.702 & 0.698 & 0.702 & 0.761 \\ 
NumHTML (w/o Pareto MTL) & 0.293 & \textbf{0.129} & 0.198 & 0.133 & 0.701 & \textbf{0.700} & 0.711 & 0.759\\
NumHTML (w/o Adaptive Pre-training) & 0.282 & 0.121 & 0.199 & 0.130 & 0.697 & 0.668 & 0.705 & 0.746\\
NumHTML (Multi-modal) & \textbf{0.325**} & 0.126 & \textbf{0.206**} & \textbf{0.136*} & \textbf{0.722*} & 0.697 & \textbf{0.716*} & \textbf{0.770**}\\ \hline
\end{tabular}
\end{table*}

\begin{table}[t]
\centering
\small
\caption{Cumulative profit across different trading strategies.}
\begin{tabular}{lll}
\hline
\textbf{Strategy} & \textbf{Profit} & \textbf{Sharpe Ratio} \\ \hline
\textbf{Simple Baselines} & & \\
Buy-all & \$36.59 & 0.76 \\
Short-sell-all &-\$36.59 & -0.77 \\
Random &-\$25.78 & -0.58 \\
\textbf{Multi-modal Methods} & & \\
MRDM  & \$38.75 & 0.81 \\
HTML & \$72.47 & 1.52 \\
NumHTML (w/o Pareto) & \$73.90 & 1.53 \\
Ensemble & \$75.73 & 1.59 \\
NumHTML & \$\textbf{77.81} &\textbf{1.62} \\ \hline
\end{tabular}
\end{table}

The stock movement predictions results are presented in Table 3, using each of the baselines and several variations of the NumHTML model for 3, 7, 15, and 30-day prediction periods. The results indicate that NumHTML using multi-modal data generally outperforms all alternative methods, including the current state-of-the-art, multi-modal Ensemble method. The NumHTML variants generate predictions with the highest MCC and F1 scores, compared with the similar multi-modal, multi-task approach of the Ensemble alternative \citep{sawhey2020mm}. This means that our single-model approach achieves statistically significant performance improvements over the Ensemble method for all cases when using multi-model versions. In addition, using text-only data, our approach also achieves some meaningful improvements in almost all settings, excluding \emph{n=15}. 

To further understand the benefits of NumHTML, in what follows, we also consider several ablation studies to determine the efficacy of different NumHTML components. 

\subsubsection{On the Utility of Structured Adaptive Pre-Training:}
In Table 3, we see NumHTML prediction performance exceeds that of NumHTML without the structured adaptive pre-training, for all \(n\). In other words, by better modeling the numerical aspects of earnings call data, it is possible to significantly improve subsequent prediction performance. 

\subsubsection{On the Utility of Volatility Prediction as an Auxiliary Prediction Task:}
Overall, NumHTML also significantly outperforms baseline methods in the volatility prediction task. Moreover, by comparing our approach with and without Pareto MTL, in Table 3, we see that the volatility prediction task consistently improves the results as an auxiliary task when predicting the stock movement; the single exception is for $n = 7$. Moreover, Figure 3 shows that NumHTML can even achieve the best average performance (least MSE error over four sub-tasks) for the auxiliary task, which is ignored in previous works \cite{yang2020html}. Thus, by comparing the volatility prediction results of our approach with and without Pareto MTL, we find that the trade-off considerations between two tasks can significantly improve the performance of the auxiliary task.

\subsubsection{On the Utility of Audio Features:}

While existing work \citep{Qin19,yang2020html,sawhey2020mm} only explores the utility of audio features for volatility prediction, Table 3 shows how multi-modal learning consistently outperforms methods, which are purely based on the textual features, for stock movement prediction also.  In particular, we observe consistent improvements by using the vocal cues compared with text-only versions among the four multi-modal methods. Improvements of this scale, relative to the corresponding text-only versions, are likely to translate into substantial practical benefits and suggest significant value in the use of audio features for a range of financial forecasting problems.

\subsection{Cumulative Profit in a Trading Simulation \label{sec:trade}}
Next, we consider the value of the various approaches to stock movement prediction in the context of a more realistic trading simulation. Table 4 presents the results (cumulative profit achieved and Sharpe Ratio) for various approaches.  We use three standard trading strategies as baseline strategies: \textit{Buy-all}, \textit{Short-sell-all}, \textit{Random} that are commonly used as benchmarks. We also compare our method with three strong multi-modal baselines, namely MRDM \citep{Qin19}, HTML \citep{yang2020html}, and Ensemble method \citep{sawhey2020mm}. Once again, the NumHTML approach outperforms all of the alternatives in terms of both profit achieved and the Sharpe Ratio (higher is better). The profit achieved by NumHTML significantly exceeds that of the S\&P 500 over the same period.

We have also interested in the individual effect of \textit{structured adaptive pre-training} and \textit{pareto multi-task learning}, respectively. Comparing the NumHTML to HTML with adaptive pre-training only (shown as NumHTML w/o Pareto), Table 4 shows that HTML with adaptive pre-training can improve the arbitrage profit somewhat, but without benefiting the Sharpe Ratio. Furthermore, the single Pareto MTL component provides significant performance benefits in terms of profit (73.90 vs 77.81) and Sharpe Ratio (1.53 vs 1.62), which suggests that our method benefits considerably from the trade-off considerations.

\begin{figure}[t]
  \centering
  \includegraphics[width = .8\linewidth]{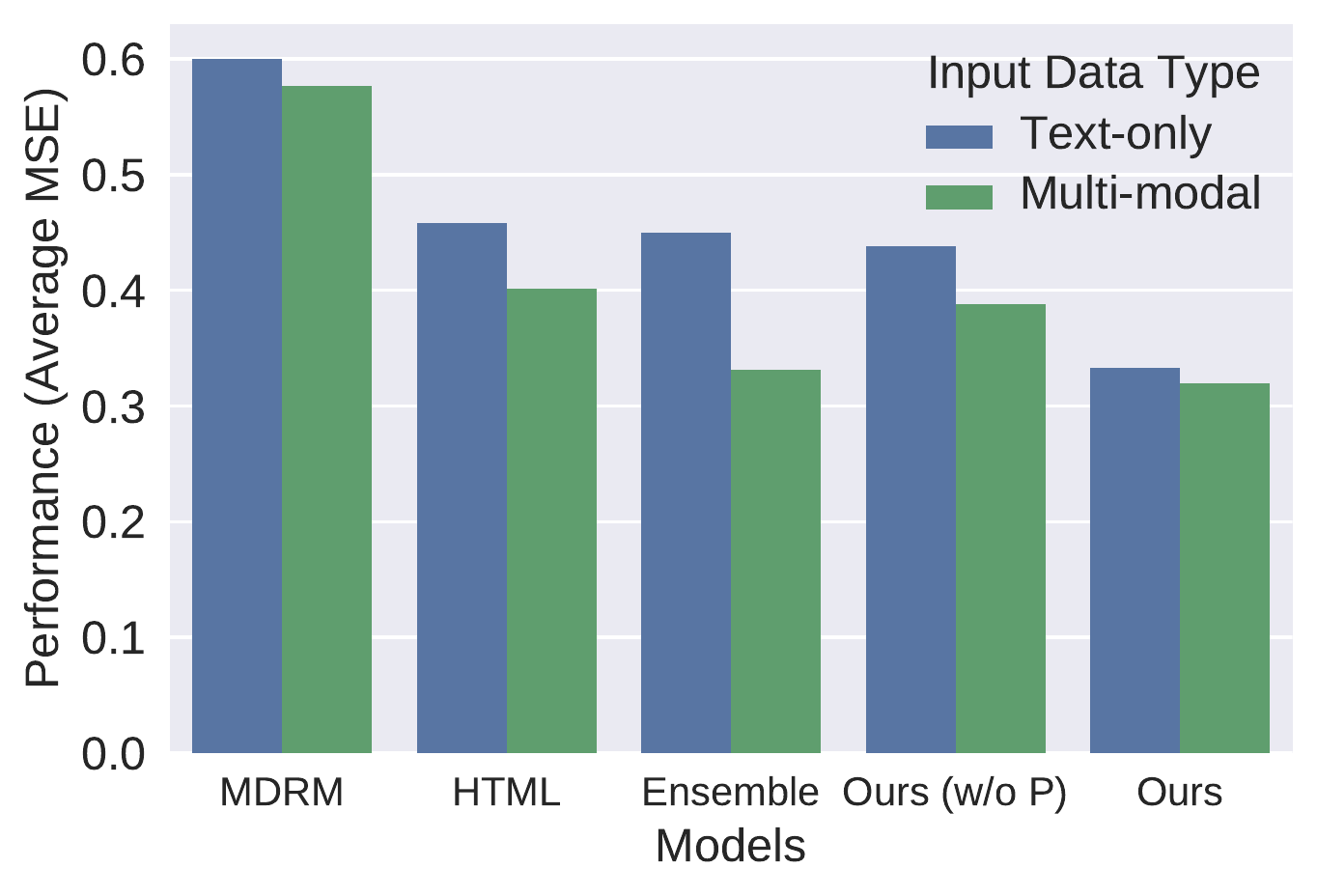}
  \caption{The results of volatility prediction. 'Ours (w/o P)' indicates NumHTML without Pareto MTL. Text-only and multi-modal methods are presented by different colors.}
  \label{fig:c63}
\end{figure}

\section{Conclusion}
This work contributes to multi-task financial forecasting, with a particular focus on the stock movement prediction, by using multi-modal earnings conference calls data. In particular, we propose a novel, numeric-oriented hierarchical transformer-based model (NumHTML) by using structured adaptive pre-training to improve how numeric data is represented and used in the pre-trained language model. A comprehensive comparative evaluation demonstrates significant performance benefits accruing to NumHTML, compared to a variety of state-of-the-art baselines and in the context of stock prediction and extended trading tasks. This evaluation also includes an ablation study to clarify the utility of different NumHTML components (adaptive pre-training, auxiliary volatility prediction, and the use of audio features). 

Our work may be extended in several ways. More sophisticated numeric representations can be imagined in order to improve the representation of numeric data. Likewise, it may be feasible to develop similar representations for other categories of useful data in due course. In this work, we focused on stock movement prediction and trading, but the approaches described may be of value in a range of financial forecasting tasks such as portfolio management/design, hedging, financial fraud or accounting errors, etc. The current trading simulation, based on \citep{sawhey2020mm}, imposes a significant single-stock purchasing limit per time period, as discussed. Going forward it will be necessary to consider more sophisticated trading policies. 

\section{Acknowledgments}
We acknowledge with thanks the discussion with Boyuan Zheng and Cunxiang Wang from Westlake University, as well as the many others who have. We would also like to thank anonymous reviewers for their insightful comments and suggestions to help improve the paper. This publication has emanated from research conducted with the financial support of Rong Hui Jin Xin Company under Grant Number 10313H041801.

\bibliography{references}

\end{document}